\documentclass[runningheads]{llncs}
\usepackage[T1]{fontenc}
\usepackage{marvosym}
\usepackage[table]{xcolor} 
\definecolor{lightgray}{RGB}{240,240,240}
\definecolor{lightred}{RGB}{255, 230, 240}
\usepackage{graphicx,verbatim}
\usepackage{makecell}
\usepackage{amsmath}
\usepackage{amssymb}
\usepackage{booktabs}
\usepackage{bbm}
\usepackage{bm}
\usepackage{enumitem}
\usepackage{pifont}
\usepackage[export]{adjustbox}
\usepackage{multirow}
\usepackage{overpic}
\usepackage{minitoc}
\usepackage[accsupp]{axessibility}
\usepackage[colorlinks, linkcolor=blue, citecolor=blue]{hyperref}
\begin{document}
\title{NEARL: Interacted Query Adaptation with Orthogonal Regularization for Medical Vision-Language Understanding}
%

\author{Zelin Peng\textsuperscript{\dag}\inst{1} \and
        Yichen Zhao\textsuperscript{\dag}\inst{1} \and
        Yu Huang\inst{1} \and
        Piao Yang\inst{2} \and
        Feilong Tang\inst{3} \and
        Zhengqin Xu\inst{4} \and
        Xiaokang Yang\inst{1} \and
        Wei Shen\textsuperscript{\Letter}\inst{1}}
\authorrunning{Z. Peng and Y. Zhao et al.}
%
\institute{MoE Key Lab of Artificial Intelligence, AI Institute, School of Computer Science, Shanghai Jiao Tong University \and
            Department of Radiology, The First Affiliated Hospital, School of Medicine, Zhejiang University \and
            Mohamed bin Zayed University of Artificial Intelligence \and
            State Key Laboratory of Infrared Physics, Shanghai Institute of Technical Physics, Chinese Academy of Science
}


  
\maketitle              
\begingroup
\renewcommand{\thefootnote}{}
\footnotetext{\textsuperscript{\dag} Equal contribution.}
\footnotetext{\textsuperscript{\Letter} Corresponding author: \texttt{wei.shen@sjtu.edu.cn}}
\addtocounter{footnote}{-2}
\endgroup
\begin{abstract}
Computer-aided medical image analysis is crucial for disease diagnosis and treatment planning. While vision-language models (VLMs) such as CLIP exhibit strong generalization ability, their direct application to medical imaging remains hindered by a substantial domain gap. Existing methods for bridging this gap, including prompt learning and unidirectional modality interaction, typically introduce domain knowledge into only one modality. However, such approaches fail to fully exploit CLIP's inherent dual-modality structure and overlook the synergistic effect of bidirectional cross-modal interaction, resulting in persistent modality misalignment. In this paper, we propose \textbf{NEARL} (i\underline{N}teracted qu\underline{E}ry \underline{A}daptation with o\underline{R}thogona\underline{L} Regularization), a novel parameter-efficient VLM framework for bidirectional cross-modal interaction. NEARL consists of two key components: (1) the Unified Synergy Embedding Transformer (USEformer), which dynamically generates compact cross-modal queries to facilitate interaction; and (2) the Orthogonal Cross-Attention Adapter (OCA), which decouples new knowledge into truly novel and incremental components through orthogonal regularization. This design reduces interference from incremental components, enabling more focused learning of novel information and improving modality interaction in VLMs. Notably, NEARL introduces only \textbf{1.46M} learnable parameters. Extensive experiments on three medical imaging modalities demonstrate state-of-the-art performance (e.g., a \textbf{2.3\%} relative improvement on the pneumonia dataset), along with fast inference and low memory overhead, highlighting its effectiveness for real-world medical vision-language understanding.

\keywords{Medical Vision-Language Adaptation \and Bidirectional Modality Interaction \and Orthogonal Feature Decoupling}

\end{abstract}
\section{Introduction}

Medical imaging analysis (e.g., disease classification) is a cornerstone of the biomedical field~\cite{xcoop_2024_miccai,pm2_2024_bibm,swinunet_2022_eccv,zhao2026thinking,multi_2022_miccai,voco_2024_cvpr,mim_2025_tmi}, enabling critical clinical applications such as disease diagnosis and treatment planning. Given the scarcity of large-scale annotated datasets for supervised learning in medical imaging, researchers are increasingly turning to Vision-Language Models (VLMs) pre-trained on vast amounts of image-text data, e.g., CLIP~\cite{clip_2021_icml}. However, their direct application to medical image analysis is impeded by a domain gap~\cite{xcoop_2024_miccai}. This gap exists because these models are often pre-trained on natural images and thus fail to account for the unique properties of medical images, e.g., their distinct anatomical structures, varied imaging modalities, and specific disease manifestations. 

To solve this issue, existing approaches can be broadly categorized into two main strategies: \noindent(\textit{i}) \textbf{Prompt learning}. These techniques~\cite{coop_2022_ijcv,xcoop_2024_miccai,vip_2024_miccai} focus on refining the prompt descriptions for the textual modality. The objective is to design prompts that are semantically aligned with the medical domain, thereby creating a better counterpart for the visual features. Nevertheless, the limited flexibility of the prompt space in a single modality often results in only modest performance gains. \noindent(\textit{ii}) \textbf{Unidirectional modality interaction}. These methods~\cite{cocoop_2022_CVPR,maple_2023_cvpr,fate_2025_aaai} facilitate cross-modal interactions in a unidirectional style.  This unidirectional flow fails to fully exploit the intrinsic duality of CLIP, allowing any error from a dominant modality to cascade and accumulate in the subordinate modality without any mechanism for correction. This cascading error effect ultimately hinders modality alignment and fails to unlock the full potential of VLMs.

In this paper, we introduce \textbf{NEARL} (i\underline{N}teracted qu\underline{E}ry \underline{A}daptation with o\underline{R}thogona\underline{L} Regularization), a novel framework to unlock the full potential of VLMs for medical vision-language understanding. NEARL is built upon two core modules: USEformer and OCA.
\noindent(\textit{i}) \textbf{Unified Synergy Embedding Transformer (USEformer}). USEformer enables bidirectional cross-modal enrichment: a set of learnable queries extracts and summarizes relevant information from one modality to produce compact and complementary representations for the other, enriching text with visual context and visuals with textual meaning, thereby achieving deep mutual alignment. \noindent(\textit{ii}) \textbf{Orthogonal Cross-Attention Adapter (OCA}). Following the cross-modal interaction in USEformer, OCA is employed to decompose the resulting features. Using Gram-Schmidt orthogonalization~\cite{schmidt1908auflosung}, OCA decouples this new knowledge into two orthogonal components: (1) truly novel, domain-specific information, and (2) incremental updates relative to the frozen pre-trained weights. This decomposition is critical as it prevents feature interference with the model's pre-trained generalization capabilities, thereby enabling a more focused acquisition of new knowledge.

We extensively evaluate NEARL on three medical imaging datasets spanning three imaging modalities and demonstrate state-of-the-art performance against existing methods. Our experiments show significant improvements in commonly used metrics, such as classification accuracy (up to \textbf{2.3\%} relative ACC gain on the pneumonia dataset~\cite{2018_cell_OCT_pneumonia_dataset}), with only \textbf{1.46M} additional learnable parameters.
\section{Methodology}

\begin{figure}[t]
    \centering
    \includegraphics[width=1.0\linewidth]{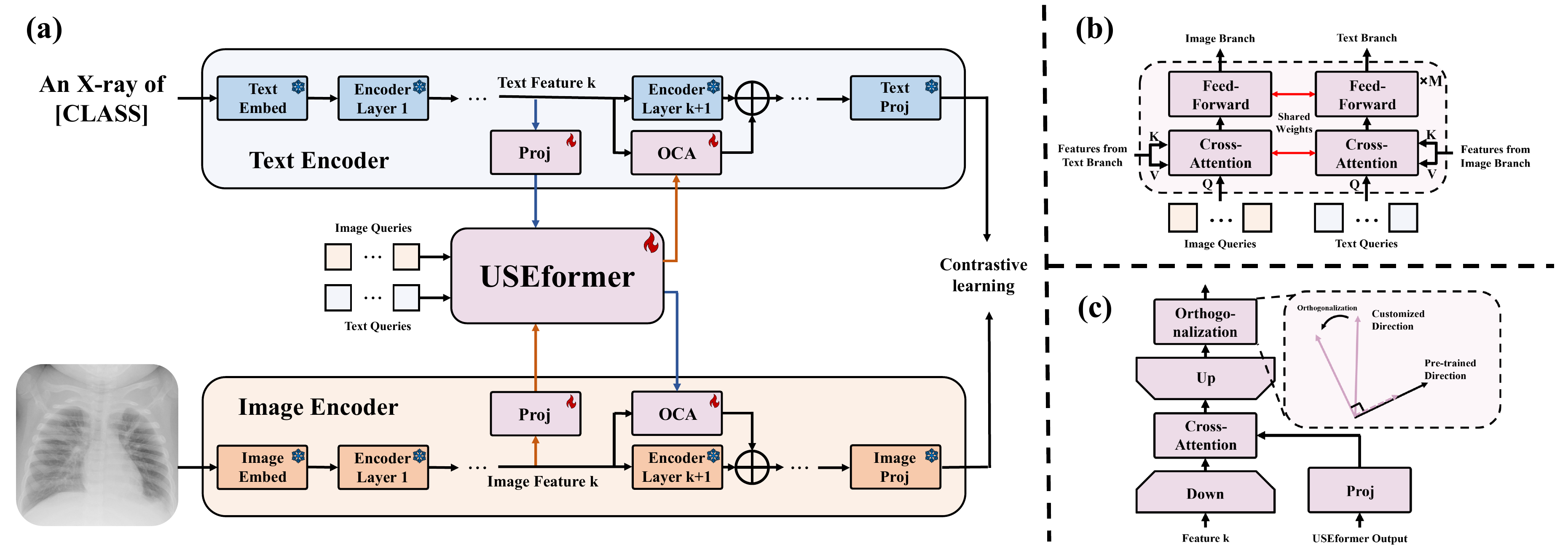}
    \caption{\textbf{Architecture of NEARL.} NEARL consists of two core components: USEformer, which enables bidirectional interaction between visual and textual features for cross-modal knowledge fusion; and OCA, which injects the cross-modal residual into the frozen pre-trained encoders through an orthogonal projection, thereby mitigating feature interference during adaptation.}
    \label{fig1}
\end{figure}

Figure~\ref{fig1} illustrates the overall architecture of our proposed NEARL. We first introduce the overall pipeline (Sec.~\ref{sec.1}) and then describe its two core components, USEformer (Sec.~\ref{sec.2}) and OCA (Sec.~\ref{sec.3}).

\subsection{Overall Pipeline}
\label{sec.1}

NEARL builds upon CLIP~\cite{clip_2021_icml} and retains its pre-trained image and text encoders. Given an input image $\mathbf{I}$ and disease classes $\mathcal{C} = \{1,\dots,C\}$, we construct textual prompts using the template ``An [modality] of [CLASS]'', yielding $\mathbf{T} = \{T_1,\dots,T_C\}$, where [modality] specifies the imaging modality and [CLASS] denotes the disease category. The image encoder $\mathcal{I}$ extracts global and local visual features, while the text encoder $\mathcal{T}$ generates class-specific textual features:
\begin{equation}
\mathbf{V} = \mathcal{I}(\mathbf{I}; \mathbf{W}^{v}, \boldsymbol{\theta}^{v}),
\end{equation}
\begin{equation}
\mathbf{S} = \mathcal{T}(\mathbf{T}; \mathbf{W}^{t}, \boldsymbol{\theta}^{t}),
\end{equation}
where $\mathbf{W}^{v}$ and $\mathbf{W}^{t}$ denote the frozen pre-trained weights of the image and text encoders, and $\boldsymbol{\theta}^{v}$ and $\boldsymbol{\theta}^{t}$ denote the trainable parameters introduced by USEformer and OCA. The visual output is denoted as $\mathbf{V} = [\mathbf{v}^{g}; \mathbf{V}^{l}]$, where $\mathbf{v}^{g} \in \mathbb{R}^{D^v}$ is the global visual feature and $\mathbf{V}^{l} \in \mathbb{R}^{N^v \times D^v}$ contains the local visual features. $\mathbf{S} = \{\mathbf{s}_{1}, \dots, \mathbf{s}_{C}\}$ denotes the textual features of all class prompts.

\noindent \textbf{Training objectives.}
Following CLIP, NEARL uses separate projectors for the image and text branches to obtain the projected image feature $\mathbf{v}$ and class-level text features $\mathbf{S} = \{\mathbf{s}_1, \mathbf{s}_2, \ldots, \mathbf{s}_C\}$ from the global image feature $\mathbf{v}^{g}$ and the end-of-sequence (EOS) token feature of each class prompt. The projectors map the two modalities into a shared embedding space, where image-text similarities are computed using dot products between normalized features. Since each image is associated with a single ground-truth disease category, NEARL is optimized using the cross-entropy loss:
\begin{equation}
\mathcal{L}_{\mathrm{ce}} = - \log \frac{\exp(\mathbf{v}^{\top} \mathbf{s}_{y} / \tau)}{\sum_{i=1}^{C} \exp(\mathbf{v}^{\top} \mathbf{s}_{i} / \tau)},
\end{equation}
where $\mathbf{s}_{y}$ denotes the text feature corresponding to the ground-truth disease category, and $\tau$ is a temperature parameter.

\subsection{Unified Synergy Embedding Transformer}
\label{sec.2}

USEformer is a lightweight module designed to facilitate bidirectional cross-modal interaction between the visual and textual branches. As shown in Figure~\ref{fig1}, USEformer receives intermediate image and text features from the frozen encoders, together with learnable image and text query tokens. It consists of $M$ stacked layers, where each layer uses cross-attention and a feed-forward network (FFN) to generate cross-modal representations conditioned on the other modality.

Given the intermediate image features $\mathbf{f}^{v}_{k} \in \mathbb{R}^{N^{v} \times D^{v}}$ and text features $\mathbf{f}^{t}_{k} \in \mathbb{R}^{N^{t} \times D^{t}}$ from the $k$-th encoder layer, we first project them into a shared low-dimensional space using separate linear projections:
\begin{equation}
\mathbf{h}^{v}_{k} = P^{v}_{k}(\mathbf{f}^{v}_{k}), 
\qquad
\mathbf{h}^{t}_{k} = P^{t}_{k}(\mathbf{f}^{t}_{k}),
\end{equation}
where $\mathbf{h}^{v}_{k} \in \mathbb{R}^{N^{v} \times D^{q}}$ and $\mathbf{h}^{t}_{k} \in \mathbb{R}^{N^{t} \times D^{q}}$. The learnable image queries $\mathbf{q}^{v} \in \mathbb{R}^{N^{q} \times D^{q}}$ and text queries $\mathbf{q}^{t} \in \mathbb{R}^{N^{q} \times D^{q}}$ then interact with the projected features through cross-attention. Specifically, visual information is injected into the textual query tokens by:
\begin{equation}
\mathrm{Attn}_{V\rightarrow T}(\mathbf{q}^{t}, \mathbf{h}^{v}_{k}) =
\operatorname{softmax}\left(
\frac{(\mathbf{q}^{t}W_Q)(\mathbf{h}^{v}_{k}W_K)^{\top}}{\sqrt{D^{q}}}
\right)(\mathbf{h}^{v}_{k}W_V),
\end{equation}
while textual information is injected into the visual query tokens by:
\begin{equation}
\mathrm{Attn}_{T\rightarrow V}(\mathbf{q}^{v}, \mathbf{h}^{t}_{k}) =
\operatorname{softmax}\left(
\frac{(\mathbf{q}^{v}W_Q)(\mathbf{h}^{t}_{k}W_K)^{\top}}{\sqrt{D^{q}}}
\right)(\mathbf{h}^{t}_{k}W_V).
\end{equation}
Here, $W_Q$, $W_K$, and $W_V$ are learnable projection matrices. For parameter efficiency, they are shared across the two cross-attention branches. The outputs of the cross-attention blocks are further processed by FFNs to obtain the image-conditioned textual representation $\mathbf{z}^{t}_{k}$ and the text-conditioned visual representation $\mathbf{z}^{v}_{k}$. By stacking this block $M$ times, USEformer iteratively refines the cross-modal representations before they are injected into the corresponding encoder layers through OCA.

\subsection{Orthogonal Cross-Attention Adapter}
\label{sec.3}

To inject the cross-modal representations produced by USEformer into the frozen CLIP encoders, we introduce OCA, a lightweight adapter module placed at the $(k+1)$-th layer of each modality branch. OCA takes two inputs: the intermediate feature from the preceding encoder layer and the cross-modal representation generated by USEformer. These two inputs are projected into a common feature space and fused via cross-attention, allowing the adapter to incorporate cross-modal context while keeping the original pre-trained encoder weights frozen.

For the image or text branch, this operation is formulated as:
\begin{equation}
\Delta \mathbf{f}^{v/t}_{k+1} =
W_{u,k+1}^{v/t} \,
\mathrm{Attn}\bigl(
W_{d,k+1}^{v/t} \mathbf{f}^{v/t}_{k},
W_{p,k+1}^{v/t} \mathbf{z}^{v/t}_{k}
\bigr),
\end{equation}
where $W_{d,k+1}^{v/t}$, $W_{p,k+1}^{v/t}$, and $W_{u,k+1}^{v/t}$ are learnable projection layers. In this formulation, the first argument of $\mathrm{Attn}(\cdot,\cdot)$ serves as the query, while the second argument provides the key and value features. The resulting $\Delta \mathbf{f}^{v/t}_{k+1}$ is an adaptive residual that carries cross-modal information for the corresponding modality branch.

\noindent \textbf{Orthogonal Projection.}
For simplicity, we omit the modality superscript \(v/t\) in the following formulation. Directly adding the adaptive residual to the pre-trained feature may interfere with the original representation learned by CLIP. To mitigate this issue, OCA introduces an orthogonal projection operation derived from the Gram-Schmidt process~\cite{schmidt1908auflosung}. Specifically, instead of directly adding the adaptive residual to the pre-trained feature, we project the residual onto the orthogonal complement of the corresponding pre-trained feature. This projection is applied token-wise:
\begin{equation}
\Delta \mathbf{f}^{\perp}_{k+1,n}
=
\Delta \mathbf{f}_{k+1,n}
-
\frac{
\langle \Delta \mathbf{f}_{k+1,n}, \mathbf{f}_{k+1,n} \rangle
}{
\langle \mathbf{f}_{k+1,n}, \mathbf{f}_{k+1,n} \rangle + \epsilon
}
\mathbf{f}_{k+1,n},
\end{equation}
where \(n\) denotes the token index, \(\langle \cdot, \cdot \rangle\) denotes the vector inner product, and \(\epsilon\) is a small constant for numerical stability. This orthogonal projection encourages the adaptive residual to encode complementary information while preserving the pre-trained representation as the main feature backbone.

The final output feature of the $(k+1)$-th layer is obtained by adding the orthogonal adaptive residual to the frozen pre-trained feature:
\begin{equation}
\tilde{\mathbf{f}}^{v/t}_{k+1}
=
\mathbf{f}^{v/t}_{k+1}
+
\Delta \mathbf{f}^{v/t,\perp}_{k+1}.
\end{equation}
The updated feature $\tilde{\mathbf{f}}^{v/t}_{k+1}$ is then passed to the subsequent encoder layers, enabling NEARL to perform cross-modal adaptation while maintaining the stability of the frozen pre-trained encoders.
\section{Experiment}

In this section, we first describe the experimental setup and implementation details (Sec.~\ref{5.1}). We then present a comparative analysis of NEARL against state-of-the-art methods (Sec.~\ref{5.2}) and evaluate its computational efficiency (Sec.~\ref{5.3}). We also conduct ablation studies (Sec.~\ref{5.4}) and perform visual analyses of the learned feature spaces to explain why fine-tuning only the language branch is insufficient in medical domains (Sec.~\ref{5.5}).

\begin{table}[t]
    \centering
    \caption{Performance comparison of state-of-the-art methods in terms of ACC (\%) and F1 (\%) on the three medical datasets. We re-implemented all compared methods using their official code. \textbf{Bold} denotes the best result, and \underline{underline} indicates the second-best.}\label{tab1}
    \fontsize{8}{10}\selectfont
    \begin{tabular}{lccccccc}
        \toprule
        \multirow{2}{*}{\textbf{Method}} & \multirow{2}{*}{\textbf{Backbone}} & \multicolumn{2}{c}{\textbf{Pneumonia}} & \multicolumn{2}{c}{\textbf{Alzheimer}} & \multicolumn{2}{c}{\textbf{Retina}} \\
        \cmidrule(lr){3-4} \cmidrule(lr){5-6} \cmidrule(lr){7-8} 
        & & \textbf{ACC} & \textbf{F1} & \textbf{ACC} & \textbf{F1} & \textbf{ACC} & \textbf{F1} \\
        \midrule
        \rowcolor{lightgray}
        \multicolumn{8}{c}{\textbf{Unimodal Methods}} \\
        ViT~\cite{ViT2021ICLR} & Transformer & 87.20\textsubscript{0.6} & 85.50\textsubscript{0.7} & 87.00\textsubscript{0.3} & 86.90\textsubscript{0.3} & 94.50\textsubscript{0.2} & 93.20\textsubscript{0.2} \\
        MedMamba~\cite{yue2024medmamba} & Mamba & 89.50\textsubscript{1.3} & 88.10\textsubscript{1.6} & \underline{90.70\textsubscript{1.7}} & \underline{90.70\textsubscript{1.7}} & 97.50\textsubscript{0.6} & 96.90\textsubscript{0.7} \\
        \midrule
        \rowcolor{lightgray}
        \multicolumn{8}{c}{\textbf{Prompt Learning Methods}} \\
        CoOp~\cite{coop_2022_ijcv} & CLIP & 83.40\textsubscript{2.8} & 81.30\textsubscript{3.7} & 85.80\textsubscript{0.0} & 85.80\textsubscript{0.1} & 94.50\textsubscript{0.1} & 93.10\textsubscript{0.3} \\
        ViP~\cite{vip_2024_miccai} & CLIP & 84.60\textsubscript{2.2} & 82.60\textsubscript{3.1} & 85.60\textsubscript{0.4} & 85.60\textsubscript{0.4} & 95.00\textsubscript{0.2} & 93.90\textsubscript{0.3} \\
        XCoOp~\cite{xcoop_2024_miccai} & CLIP & 88.10\textsubscript{1.5} & 86.60\textsubscript{1.8} & 90.40\textsubscript{0.4} & 90.40\textsubscript{0.4} & \underline{97.70\textsubscript{0.1}} & \underline{97.30\textsubscript{0.0}} \\
        \midrule
        \rowcolor{lightgray}
        \multicolumn{8}{c}{\textbf{Unidirectional Modality Interaction Methods}} \\
        CoCoOp~\cite{cocoop_2022_CVPR} & CLIP & 82.60\textsubscript{0.9} & 80.00\textsubscript{1.1} & 85.60\textsubscript{0.2} & 85.50\textsubscript{0.2} & 95.00\textsubscript{0.4} & 93.90\textsubscript{0.5} \\
        MaPLe~\cite{maple_2023_cvpr} & CLIP & \underline{92.60\textsubscript{3.2}} & \underline{91.70\textsubscript{3.8}} & \underline{90.70\textsubscript{1.2}} & 90.60\textsubscript{0.2} & 97.60\textsubscript{0.4} & 97.10\textsubscript{0.5} \\
        FATE~\cite{fate_2025_aaai} & CLIP & 82.60\textsubscript{0.7} & 80.40\textsubscript{0.9} & 84.10\textsubscript{0.8} & 84.00\textsubscript{0.8} & 92.30\textsubscript{0.5} & 90.90\textsubscript{0.6} \\
        TextRefiner~\cite{textrefiner_2025_aaai} & CLIP & 84.00\textsubscript{0.3} & 81.80\textsubscript{0.5} & 85.10\textsubscript{0.5} & 85.00\textsubscript{0.5} & 94.10\textsubscript{0.3} & 92.70\textsubscript{0.4} \\
        \midrule
        \rowcolor{lightgray}
        \multicolumn{8}{c}{\textbf{Bidirectional Modality Interaction Methods}} \\
        \textbf{NEARL} \textsubscript{ours} & \textbf{CLIP} & \textbf{94.70\textsubscript{0.2}} & \textbf{94.20\textsubscript{0.3}} & \textbf{92.60\textsubscript{0.6}} & \textbf{92.60\textsubscript{0.6}} & \textbf{98.50\textsubscript{0.2}} & \textbf{98.20\textsubscript{0.2}} \\
        \bottomrule
    \end{tabular}%
\end{table}

\subsection{Experimental Settings}
\label{5.1} 
\noindent \textbf{Dataset.} We comprehensively evaluate the proposed method on three public medical datasets: Pneumonia~\cite{2018_cell_OCT_pneumonia_dataset}, Alzheimer~\cite{2007_brain_dataset}, and Retina~\cite{2018_cell_OCT_pneumonia_dataset}. These datasets vary in two key aspects: (\textit{i}) lesion sites, covering pneumonia, Alzheimer's disease, and diabetic macular edema; and (\textit{ii}) imaging modalities, including X-ray, MRI, and OCT. \noindent \textbf{Pneumonia} contains 5,856 chest radiograph images categorized as normal or pneumonia. We split the data according to the division protocol of~\cite{vip_2024_miccai}, resulting in a 4,710/522/624 split for training, validation, and testing. \noindent \textbf{Alzheimer} comprises 80,000 brain MR images across four stages: non-demented, very mild dementia, mild dementia, and moderate dementia. To balance positive and negative samples, we select mild and moderate dementia cases together with a random subset of non-demented images, resulting in a 7,033/1,758/2,199 split for training, validation, and testing. \noindent \textbf{Retina} contains 76,607 OCT images that cover four conditions: choroidal neovascularization (CNV), diabetic macular edema (DME), drusen, and normal eyes. We select DME and normal samples for evaluation. The final dataset is partitioned into 11,887 training, 1,857 validation, and 1,859 test images.

\noindent \textbf{Implementation Details.}
We adopt ViT-B/16 as the visual backbone and set the text embedding dimension to 512, following the original CLIP setting. We use natural-domain CLIP mainly for fair comparison with prior medical CLIP adaptation methods~\cite{xcoop_2024_miccai,vip_2024_miccai}. All results are averaged over three random seeds. The same USEformer is shared across encoder layers and is repeatedly used to produce cross-modal representations from layer-specific image and text features. OCA is inserted into each layer of both the image and text encoders to inject the corresponding cross-modal residuals. The model is trained on a single NVIDIA RTX 3090 GPU for 50 epochs. We set the number of cross-attention blocks in USEformer $M$ to 6, the number of query tokens $N^q$ to 32, the query token dimension $D^q$ to 128, the adapter rank to 8, and the temperature parameter $\tau$ to $1\times10^{-2}$.

\noindent \textbf{Evaluation Metrics.}
We use Accuracy (ACC) and Macro F1-score (F1) to evaluate classification performance on the test set of the three benchmarks.

\begin{table}[t]
    \centering
    \caption{Ablation study on key modules of NEARL.}\label{tab2}
    \fontsize{9}{10}\selectfont
    \setlength{\tabcolsep}{9pt}
    \begin{tabular}{lcccc}
        \toprule
        \textbf{Methods} & \textbf{USEformer} & \textbf{OR} & \textbf{ACC} & \textbf{F1} \\
        \midrule
        LoRA~\cite{lora_2022_iclr} & - & - & 93.3 & 92.6 \\
        \midrule
        \multirow{3}{*}{NEARL} & w/o using & $\checkmark$ & 93.7 & 93.0 \\
                               & $\checkmark$ & w/o using & 93.8 & 93.1 \\
                               & $\checkmark$ & $\checkmark$ & \textbf{94.7} & \textbf{94.2} \\
        \bottomrule
    \end{tabular}%
\end{table}

\subsection{Comparisons with the State-of-the-Arts}
\label{5.2}

We compare NEARL with state-of-the-art CLIP-based and non-CLIP-based models on three medical datasets covering X-ray, MRI, and OCT modalities, as shown in Table~\ref{tab1}. The compared CLIP-based methods include prompt-learning approaches, i.e., ViP~\cite{vip_2024_miccai}, CoOp~\cite{coop_2022_ijcv}, and XCoOp~\cite{xcoop_2024_miccai}, as well as methods based on modality interaction, i.e., CoCoOp~\cite{cocoop_2022_CVPR}, MaPLe~\cite{maple_2023_cvpr}, FATE~\cite{fate_2025_aaai}, and TextRefiner~\cite{textrefiner_2025_aaai}. We also include two non-CLIP-based unimodal models, ViT~\cite{ViT2021ICLR} and MedMamba~\cite{yue2024medmamba}. Among these methods, MedMamba, ViP, and XCoOp are specifically designed for medical image classification.

Overall, NEARL achieves the best performance across all three datasets, with accuracies of 94.7\%, 92.6\%, and 98.5\%, and F1-scores of 94.2\%, 92.6\%, and 98.2\%, respectively. These results show that NEARL generalizes effectively across different medical imaging modalities and classification tasks.

\subsection{Evaluation of Real-Time Clinical Application}
\label{5.3}

On a single NVIDIA RTX 3090 GPU with a batch size of 100, NEARL achieves 500 FPS at 1170 GFLOPs per batch, using only 1.46M trainable parameters while achieving the best accuracy, compared with existing PEFT baselines that use 8.19K to 26.11M parameters and run at 470 to 540 FPS. These results suggest that NEARL has low computational overhead and efficient inference performance, supporting its potential use in rapid clinical screening scenarios.

\subsection{Ablation Study}
\label{5.4}

\noindent We conduct ablation studies on the Pneumonia dataset. The results are reported in Table~\ref{tab2}. Removing USEformer leads to a clear performance drop, indicating that the shared cross-modal transformer is important for effective interaction between the image and text encoders. Compared with a simple linear projection, USEformer can better aggregate modality-specific features through learned cross-attention, thereby providing more informative cross-modal representations for subsequent adaptation. Removing orthogonal regularization (OR) also degrades performance, suggesting that explicitly separating the adaptive residual from the frozen pre-trained representation helps reduce feature interference during adaptation. When both USEformer and OR are removed, the framework degenerates into a LoRA-style adaptation baseline~\cite{lora_2022_iclr}, which achieves the lowest performance. This result indicates that both cross-modal interaction and orthogonal residual adaptation are necessary for the effectiveness of NEARL.

\begin{figure*}[t]
    \centering
    \includegraphics[width=\textwidth]{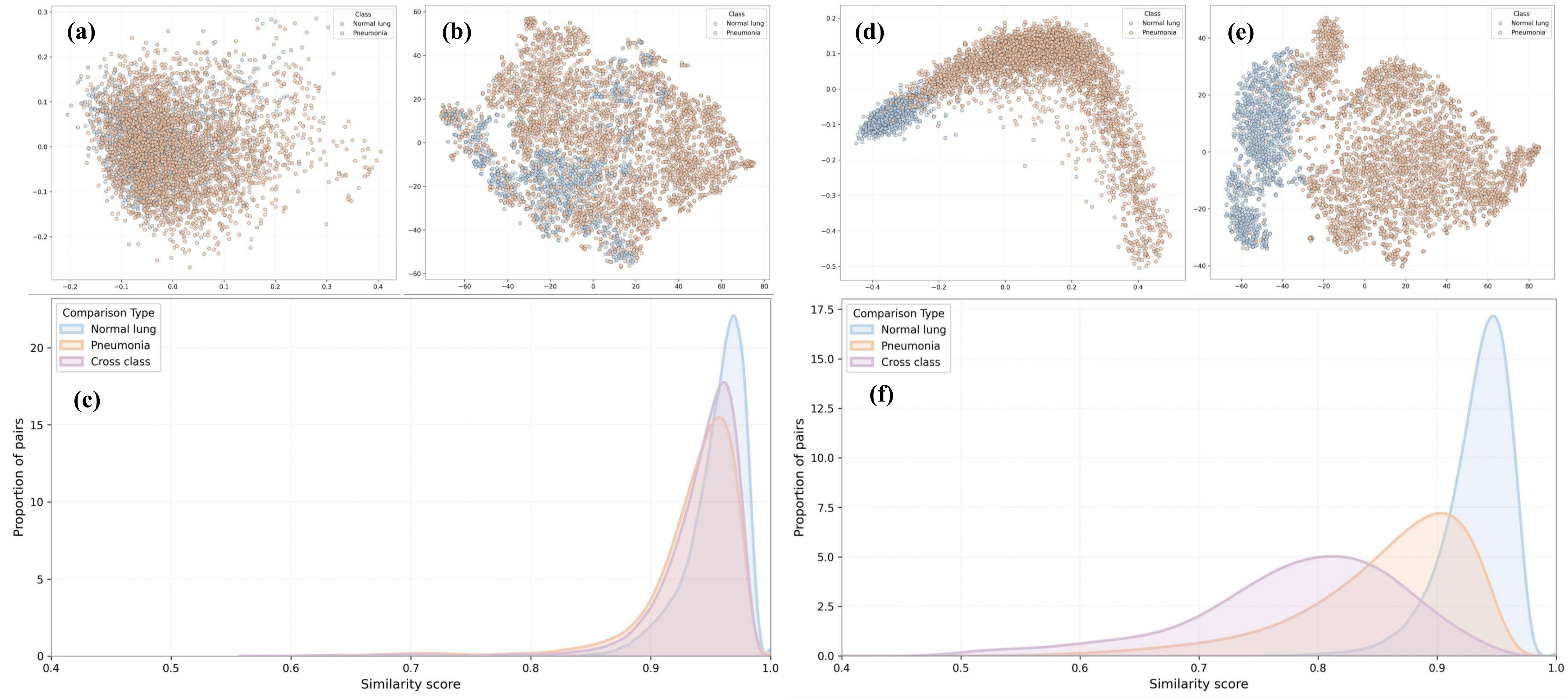}
    \caption{Visualization of CLIP (a–-c) vs. NEARL (d–-f) image features on the Pneumonia dataset.  PCA (a,d), t-SNE (b,e), and cosine similarity scores (c,f) show a large domain gap between pre-trained CLIP and medical images, leading to suboptimal generalization when only fine-tuning the text encoder. NEARL bridges this gap via bidirectional modality interaction.}
    \label{domain}
\end{figure*}

\subsection{Visual Analysis of Learned Feature Spaces}
\label{5.5}

Existing prompt-learning methods~\cite{vip_2024_miccai} typically adapt the language branch while keeping CLIP's image encoder frozen, which may be insufficient for medical image classification due to the domain gap between natural and medical images. To examine this issue, we visualize the learned feature spaces using PCA~\cite{pca} and t-SNE~\cite{tsne}. As shown in Fig.~\ref{domain}(a)--(c), pneumonia and normal chest X-ray features extracted by the frozen CLIP image encoder are highly overlapped, with weak intra-/inter-class cosine divergence, indicating limited class separability. This partly explains the limited performance of methods that keep the image encoder frozen, such as FATE~\cite{fate_2025_aaai} and TextRefiner~\cite{textrefiner_2025_aaai}. In contrast, NEARL jointly adapts the image and text branches through bidirectional cross-modal interaction, producing more separable feature distributions and clearer similarity patterns in Fig.~\ref{domain}(d)--(f). These results suggest that USEformer and OCA help learn medical-domain-aware representations while preserving the stability of the pre-trained CLIP backbone.

\section{Conclusion}

In this work, we propose \textbf{NEARL}, a parameter-efficient bidirectional framework for enhancing CLIP's cross-modal alignment in medical vision-language adaptation. By keeping CLIP's original parameters frozen and optimizing only \textbf{1.46M} additional trainable parameters, NEARL achieves substantial performance improvements with minimal computational overhead. These results suggest that NEARL provides an efficient and scalable solution for adapting vision-language foundation models to diverse medical imaging scenarios.

{
\bibliographystyle{splncs04}
\bibliography{NEARL}
}

\end{document}